\documentclass[final]{cvpr}

\usepackage{times}
\usepackage{epsfig}
\usepackage{graphicx}
\usepackage{amsmath}
\usepackage{amssymb}
\usepackage[pagebackref=true,breaklinks=true,colorlinks,bookmarks=false]{hyperref}

\def\cvprPaperID{4504} % *** Enter the CVPR Paper ID here
\def\confYear{CVPR 2021}

\begin{document}

%%%%%%%%% TITLE
\title{Watching You: Global-guided Reciprocal Learning for Video-based Person Re-identification}

%\author[1]{Xuehu Liu\thanks{snowtiger@mail.dlut.edu.cn}}

\author{{Xuehu Liu}$^{1,2}$, {Pingping Zhang}$^{1,3,*}$, {Chenyang Yu}$^{1,2}$, {Huchuan Lu}$^{1,2,4,}$\thanks{Corresponding Authors}, {Xiaoyun Yang}$^{5}$
\\
$^1$School of Information and Communication Engineering, Dalian University of Technology
\\
$^2$Ningbo Institute, Dalian University of Technology
\\
$^3$School of Artificial Intelligence, Dalian University of Technology
\\
$^4$Pengcheng Lab, $^5$Remark Holdings
\\
{\tt\small \{snowtiger, yuchenyang\}@mail.dlut.edu.cn},
%\and
%{Pingping Zhang}^{1,2}\\
{\tt\small \{zhpp, lhchuan\}@dlut.edu.cn},
{\tt\small xyang@remarkholdings.com}
}

\maketitle

%%%%%%%%% ABSTRACT
\begin{abstract}
Video-based person re-identification (Re-ID) aims to automatically retrieve video sequences of the same person under non-overlapping cameras.
To achieve this goal, it is the key to fully utilize abundant spatial and temporal cues in videos.
Existing methods usually focus on the most conspicuous image regions, thus they may easily miss out fine-grained clues due to the person varieties in image sequences.
To address above issues, in this paper, we propose a novel \textbf{Global-guided Reciprocal Learning (GRL)} framework for video-based person Re-ID.
Specifically, we first propose a \textbf{Global-guided Correlation Estimation (GCE)} to generate feature correlation maps of local features and global features, which help to localize the high- and low-correlation regions for identifying the same person.
After that, the discriminative features are disentangled into high-correlation features and low-correlation features under the guidance of the global representations.
%
%For the disentangled diverse features, we adopt distinctive strategies to fully exploit the abundant spatial-temporal information in a video.
%
Moreover, a novel \textbf{Temporal Reciprocal Learning (TRL)} mechanism is designed to sequentially enhance the high-correlation semantic information and accumulate the low-correlation sub-critical clues.
%
%Based on above modules, our approach can not only extract meaningful semantic regional features in spatial, but also keep feature consistency in temporal.
%
Extensive experiments are conducted on three public benchmarks.
The experimental results indicate that our approach can achieve better performance than other state-of-the-art approaches.
The code is released at https://github.com/flysnowtiger/GRL.
\end{abstract}

%%%%%%%%% BODY TEXT
\section{Introduction}
Person re-identification (Re-ID) aims to retrieve specific pedestrians cross different cameras at different times and places.
Recently, this task has become a hot research topic due to its importance in advanced applications, such as safe community, intelligent surveillance and criminal investigation.
%
%In the past decade, the researches have made a great progress in this filed, and derive some interesting tasks associated with person Re-ID, such as image-based person Re-ID, video-based person Re-ID, person search and so on.
%
Compared with other related Re-ID tasks, video-based person Re-ID provides a video as the input to retrieve rather than a single image.
Although videos can provide comprehensive appearance information, motion cues, pose variations in temporal, at the same time, they bring more illumination changes, complicated backgrounds and person occlusions in a clip.
Thus, there are still many challenges for researches to handle in video-based person Re-ID.

Previous methods~\cite{xu2017jointly,wu2016deep,liu2015spatio} can be coarsely summarized into two steps: spatial feature extraction and temporal feature aggregation.
First, Convolutional Neural Networks (CNNs) are utilized to extract frame-level spatial features from each single image.
Then, frame-level spatial features are temporally aggregated into a feature vector as the video representation to compute the similarity scores.
Naturally, how to fully explore the discriminative spatial-temporal cues from multiple frames is seen as the key to tackle video-based person Re-ID.
%
%Compared with a still image, the multiple frames of a video sequence usually contain comprehensive but misaligned visual information.
%
%Technically, the video representation, obtained by Global Average Pooling (GAP) followed by Temporal Average Pooling (TAP), can directly focus on the important information of persons in the sequence.
%
Generally speaking, the average pooling for spatial-temporal features can directly focus on main targets, but it has some obvious drawbacks, such as the inability to tackle the misalignment in temporal, the pollution of background noises, and the difficulty of capturing small but meaningful subjects in videos.
To address these drawbacks, in recent years, researchers have proposed some rigid-partition-based methods or soft-attention-based methods to instead the direct average operation.
These methods are beneficial to learn more discriminative and diverse local features, resulting in higher performance of video-based person Re-ID.
However, previous methods generally ignore the role of the global features in whole person recognition while strengthening the local features.
%
%For examples, Li \emph{et al.}~\cite{li2018diversity} do not consider the relationships between global and local features, and lacks global guidance for the local feature learning on each frame.
%
Based on this consideration, Zhang \emph{et al.}~\cite{zhang2020multi} utilize the local affinities with respect to inference global features to help assign different weights to local features.
Although effective, it tends to ignore inconspicuous yet fine-grained clues.
Different from~\cite{zhang2020multi}, we correlate the global feature with the pixel-level local features in a frame to generate two correlation maps, which are utilized to disentangle generic features into high- and low-correlation features.
%
%High- or low-correlation areas are adaptively mined according to the learned correlation maps.
%
%Intuitively, the higher the correlation values are, the more stable in temporal and more important in spatial the visual cues are.
%
%If the correlation values are low, the visual cues may be not continuous in temporal.
%
Intuitively, features with high correlation mean they appear frequently in temporal and are spatially conspicuous.
Features with low correlation mean they are inconspicuous and discontinuous yet meaningful.
We further explore suitable strategies for disentangled features in temporal and fully mine fine-grained cues.

Based on above considerations, we proposed a novel Global-guided Reciprocal Learning (GRL) framework for video-based person Re-ID.
The whole framework mainly consists of two key modules.
To begin with, we proposed a Global-guided Correlation Estimation (GCE) module to estimate the correlation values of frame-level local features under the global guidance.
With GCE, each frame-level feature map will be disentangled into two kinds of discriminative features with distinct correlation degrees.
The one with high correlation, usually covers the most conspicuous and continuous visual information.
Another with inverse correlation, as the supplement, is exploited to mine the fine-grained and sub-critical cues.
Besides, we propose a novel Temporal Reciprocal Learning (TRL) module to fully exploit all the discriminative features in the forward and backward process.
More specifically, for high-correlation features, we adopt a semantic enhancement strategy to mine spatial conspicuous and temporal aligned information.
For low-correlation features, we introduce a temporal memory strategy to accumulate the discontinuous but discriminative cues frame by frame.
In this way, our proposed method can not only explore the most conspicuous information from the high-correlation regions in a sequence, but also capture the sub-critical information from the low-correlation regions.
%regarded as interference information easily.
%
Extensive experiments on public benchmarks demonstrate that our framework delivers better results than other state-of-the-art approaches.

In summary, our contributions are four folds:
\begin{itemize}
\item
\vspace{-1mm}
We propose a novel Global-guided Reciprocal Learning (GRL) framework for video-based person Re-ID.
\item
\vspace{-1mm}
We propose a Global-guided Correlation Estimation module to generate the correlation maps under the guidance of video representations for disentanglement.
\item
\vspace{-1mm}
We introduce a Temporal Reciprocal Learning (TRL) module to effectively capture the conspicuous information and the fine-grained clues in videos.
\item
\vspace{-1mm}
Extensive experiments on public benchmarks demonstrate that our framework synthetically attains a better performance than several state-of-the-art methods.
\end{itemize}
%-------------------------------------------------------------------------
\section{Related Works}

\subsection{Video-based Person Re-identification}
In recent years, with the rise of deep learning~\cite{deng2009imagenet,ioffe2015batch}, person Re-ID has gained a great success and the performance has been improved significantly.
At the early stage of person Re-ID, researchers pay more attention to image-based person Re-ID.
Recently, video-based person Re-ID is seen as a generalization of image-based person Re-ID task, and has drawn more and more researchers' interests.
Generally, videos contain richer spatial and temporal information than still images.
Thus, on the one hand, some existing methods~\cite{li2018diversity,si2018dual,song2017region, fu2019sta} concentrate on extracting attentive spatial features.
On the other hand, some works~\cite{wang2014person,mclaughlin2016recurrent, li2019multi} attempt to capture temporal information to strength the video representations.
%
%Thus, how to fully utilize the spatial and temporal information is the key to tackle video-based person Re-id.
%
%Nowadays, lots of approaches~\cite{} have been proposed.
%
For example, Li \emph{et al.}~\cite{li2018diversity} employ a diverse set of spatial attention modules to consistently extract similar local patches across multiple images.
Fu \emph{et al.}~\cite{fu2019sta} design an attention module to weight horizontal parts using a spatial-temporal map for more robust clip-level feature representations .
Zhao \emph{et al.}~\cite{zhao2019attribute} propose a attribute-driven method for feature disentangling to learn various attribute-aware features.
Liu \emph{et al.}~\cite{liu2019jointpyramid} propose a soft-parsing attention network and joint utilize a spatial pyramid non-local block to learn multiple semantic-aware aligned video representations.
Zhang \emph{et al.}~\cite{zhang2020multi} utilize a representative set of reference feature nodes for modeling the global relations and capturing the multi-granularity level semantics.
In this paper, we attempt to estimate the correlation values of spatial features guided by the whole video representation, which is beneficial to cover the conspicuous visual cues in each frame.
Besides, a novel temporal reciprocal learning mechanism is proposed to explore more discriminative information for video-based person Re-ID.
%
%\subsection{Global-guided Attention}
\begin{figure*}
\centering
\resizebox{1.0\textwidth}{!}
{
\begin{tabular}{@{}c@{}c@{}}
\includegraphics[width=0.9\linewidth,height=0.55\linewidth]{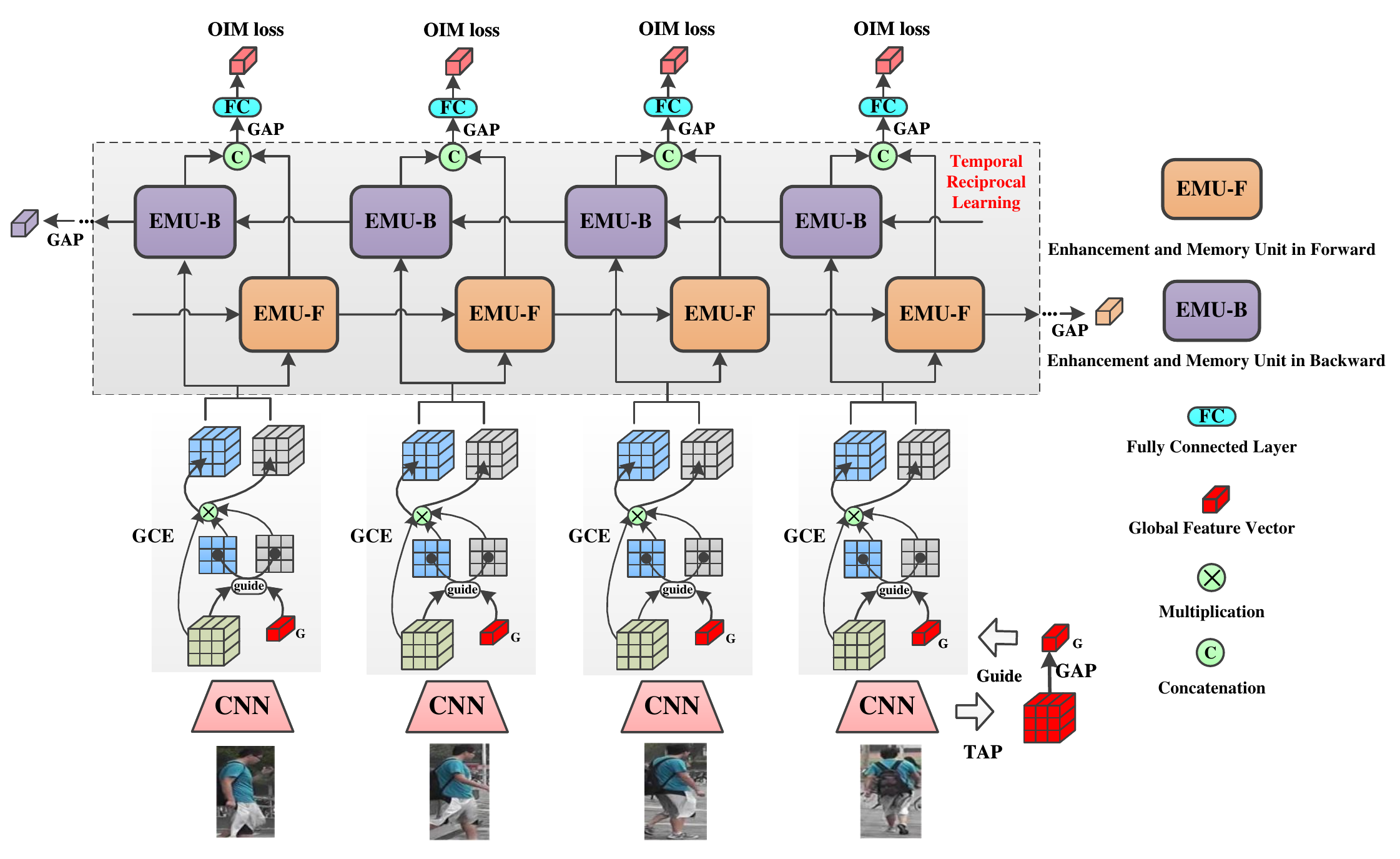} \\
\end{tabular}
}
\vspace{-4mm}
\caption{The overall structure of our proposed method.
Given an image sequence, we firstly utilize ResNet-50 to extract frame-level feature maps.
Then, frame-level features are aggregated by TAP and GAP to generate a video-level feature.
With the guidance of video-level features, Global Correlation Estimation (GCE) is utilized to generate the correlation maps for disentanglement.
Afterwards, the Temporal Reciprocating Learning (TRL) is introduced to enhance and accumulate disentangled features in forward and backward directions.
}
\label{fig:Framework}
\vspace{-2mm}
\end{figure*}
%-------------------------------------------------------------
\subsection{Temporal Feature Learning}
For video-related tasks, such as video-based person Re-ID, action recognition, video segmentation~\cite{miao2020memory} and so on, the temporal feature learning is seen as the core module in most algorithms.
Typically, the temporal information modeling methods encode temporal relations or utilize temporal cues for video representation learning.
Most of existing video-based Re-ID methods exploit optical flow~\cite{chung2017two, chen2018video}, Recurrent Neural Networks (RNN)~\cite{mclaughlin2016recurrent, hochreiter1997long}, or temporal pooling~\cite{zheng2016mars} for temporal feature learning.
For the action recognition, Weng ~\emph{et al.}~\cite{weng2020temporal}
introduce a progressive enhancement module to sequentially excite the discriminative channels of frames.
For video-based person Re-ID task, Mclaughlin \emph{et al.}~\cite{mclaughlin2016recurrent} introduce a recurrent architecture to pass the feature message of each frame for aggregating temporal information.
Xu \emph{et al.}~\cite{xu2017jointly} propose a joint spatial CNN and temporal RNN model for video-based person Re-ID.
Zhang \emph{et al.} ~\cite{zhang2018multi} introduce a reinforcement learning method for pairwise decision making.
Dai \emph{et al.}~\cite{dai2019video} design a temporal residual learning module to simultaneously extract the generic and specific features from consecutive frames.
Liu \emph{et al.}~\cite{liu2019spatial} design a refining recurrent unit and spatial-temporal integration module to integrate abundant spatial-temporal information.
Compared with existing methods, our method adopts temporal reciprocal learning for bi-directional semantic feature enhancement and temporal information accumulation.
Thus, the global-guided spatial features could focus on complementary objects, such as moving human body and key accessories.
%---------------------------------------------------------
\section{Proposed Method}
In this section, we introduce the proposed Global-guided Reciprocal Learning (GRL) framework.
We first give an overview of the proposed GRL.
Then, we elaborate the key modules in the following subsections.
\subsection{Overview}
%, which extract conspicuous information and aggregate temporal cues in a progressive manner..
The overall architecture of our proposed GRL is shown in Fig.~\ref{fig:Framework}.
Our approach consists of frame-level feature extraction, global-guided feature disentanglement, temporal reciprocal learning.
Given a video, we first use the Restricted Random Sampling (RRS)~\cite{li2018diversity} to generate training image frames.
Then, we extract frame-level features by a pre-trained backbone network (ResNet-50~\cite{he2016deep} in our work).
After that, we adopt a Temporal Average Pooling (TAP) and a Global Average Pooling (GAP) to generate a video-level representation.
With the guidance of the video-level representation, we design a Global-guided Correlation Estimation (GCE) to generate the correlation maps and disentangle the frame-level features to high- and low-correlation features. % according to the correlation degree.
Afterwards, the Temporal Reciprocating Learning (TRL) is introduced to enhance and accumulate disentangled features in forward and backward directions.
Finally, we introduce the Online Instance Matching (OIM)~\cite{xiao2017joint} loss and verification loss to optimize the whole network.
By the GRL, our method can not only capture the conspicuous information but also mine meaningful fine-grained cues in sequences.
In the test stage, the attentive pooled feature from the enhanced high-correlation vectors at different time steps and the accumulated low-correlation feature at the last time step are concatenated for the retrieval list.
%------------------------------------------------
\begin{figure}
\centering
\resizebox{0.5\textwidth}{!}
{
\begin{tabular}{@{}c@{}c@{}}
\includegraphics[width=0.5\linewidth,height=0.20\linewidth]{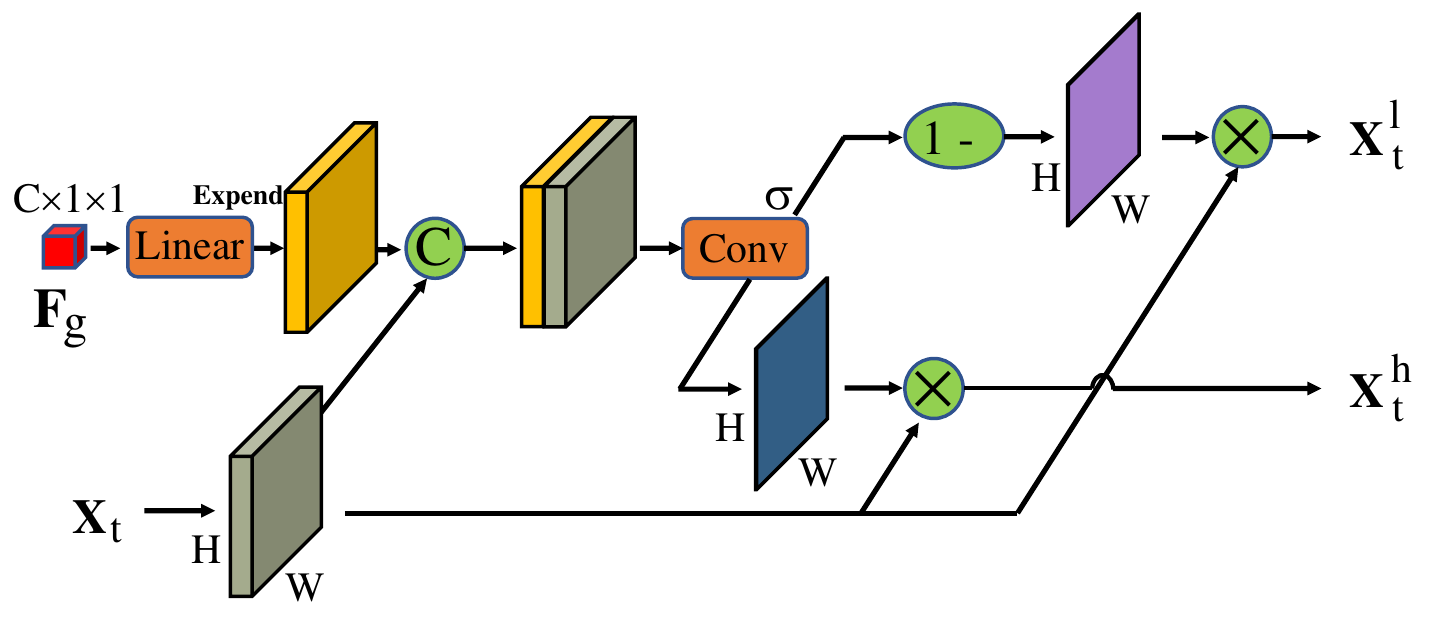} \\
\end{tabular}
}
\vspace{-4mm}
\caption{The proposed GCE module.
}
\label{fig:Global-guided}
\vspace{-4mm}
\end{figure}
%-------------------------------------------------------------
\subsection{Global-guided Feature Disentanglement}
The attention mechanism has been widely adopted to tackle the misalignment in video-based person Re-ID.
However, existing attention-based methods lack global perceptions of the whole video and easily miss out fine-grained clues.
To relieve this issue, we propose the GCE module to disentangle the spatial features into two complementary features.
One of them highlights the conspicuous information in each frame, another keeps the fine-grained and sub-critical cues.
Fig.~\ref{fig:Global-guided} shows the structure of the proposed GCE module.
Formally, given a video, we
firstly sample $T$ frames $\nu=\{\textbf{I}_1, \textbf{I}_2, ..., \textbf{I}_T \}$ as the inputs of our network.
The ResNet-50 is utilized as the feature extractor to obtain a set of frame-level features $\chi=\{\textbf{X}_t|,t=1,2,...,T\}$, where $\textbf{X}_t\in R^{C\times H\times W}$,
$H, W, C$ represents the height, width and the number of channels, respectively.
Then, we utilize a TAP and a GAP to obtain the video-level representation
\begin{equation}\label{Global feature}
\textbf{F}^g =\frac{\sum_{t,h,w=0}^{T,H,W}{X_{t,h,w}}}{H\times W \times T}
\end{equation}
$\textbf{F}^g \in R^{C\times 1 \times 1}$ can coarsely represent the whole video.

To obtain global information, the proposed GCE takes both the frame-level features $\textbf{X}_t$ and the video-level feature vector $\textbf{F}^g$ as inputs.
To guide the feature learning, the $\textbf{F}^g$ is adhered with a linear projection and expanded to $\widetilde{\textbf{F}}^g$, which has the same sizes to $\textbf{X}_t$.
The expanded features are concatenated with $\textbf{X}_t$.
Then, we integrate the global and local features, and jointly infer the degree of correlations.
%
%Though this, the each local feature on spatial is aggregated with the global feature.
%
%Concretely, two $1\times 1$ convolution layers are utilized to.
%
The correlation map $\textbf{R}_t \in R^{1\times H\times W}$ related to $\textbf{X}_t$ under the global guidance can be computed by
\begin{equation}\label{correlation map}
\textbf{R}_t = \sigma{(\textbf{W}_r([\widetilde{\textbf{F}}^g, \textbf{X}_t]))}
\end{equation}
where $[\cdot, \cdot]$ represents the concatenation operation.
$\textbf{W}_r$ is learnable weight of two 1$\times$1 convolutional layers inserted by Batch Normalization (BN) and ReLU activation.
$\sigma$ represents the sigmoid activation function.
By reversing the obtained correlation map, we can obtained a low-correlation map.
Then, the correlation maps are multiplied with original frame-level features $\textbf{X}_t$ to activate distinct local regions.
Finally, under the guidance of global representation, we disentangle frame-level features into the high-correlation features ${\textbf{X}}_t^{h}$ and the low-correlation features $\textbf{X}_t^{l}$ by
\vspace{-1mm}
\begin{equation}\label{corr applacation}
{\textbf{X}}_t^{h} = \textbf{X}_t \odot \textbf{R}_t
\end{equation}
\begin{equation}\label{uncorr applacation}
{\textbf{X}}_t^{l} = \textbf{X}_t \odot (1 - \textbf{R}_t)
\end{equation}
where $\odot$ represents element-wise multiplication, and ${\textbf{X}}_t^{h}$, $\textbf{X}_t^{l} \in R^{C\times H\times W}$.
Based on above procedures, we disentangle generic features into two distinct features, which are different from previous methods on local feature extraction.
%
%Leveraging the inverse operation on the obtained correlation map can directly suppress the highest correlation areas and strength the sub-critical regions.
%------------------------------------------------
\begin{figure}
\centering
\resizebox{0.5\textwidth}{!}
{
\begin{tabular}{@{}c@{}c@{}}
\includegraphics[width=0.45\linewidth,height=0.32\linewidth]{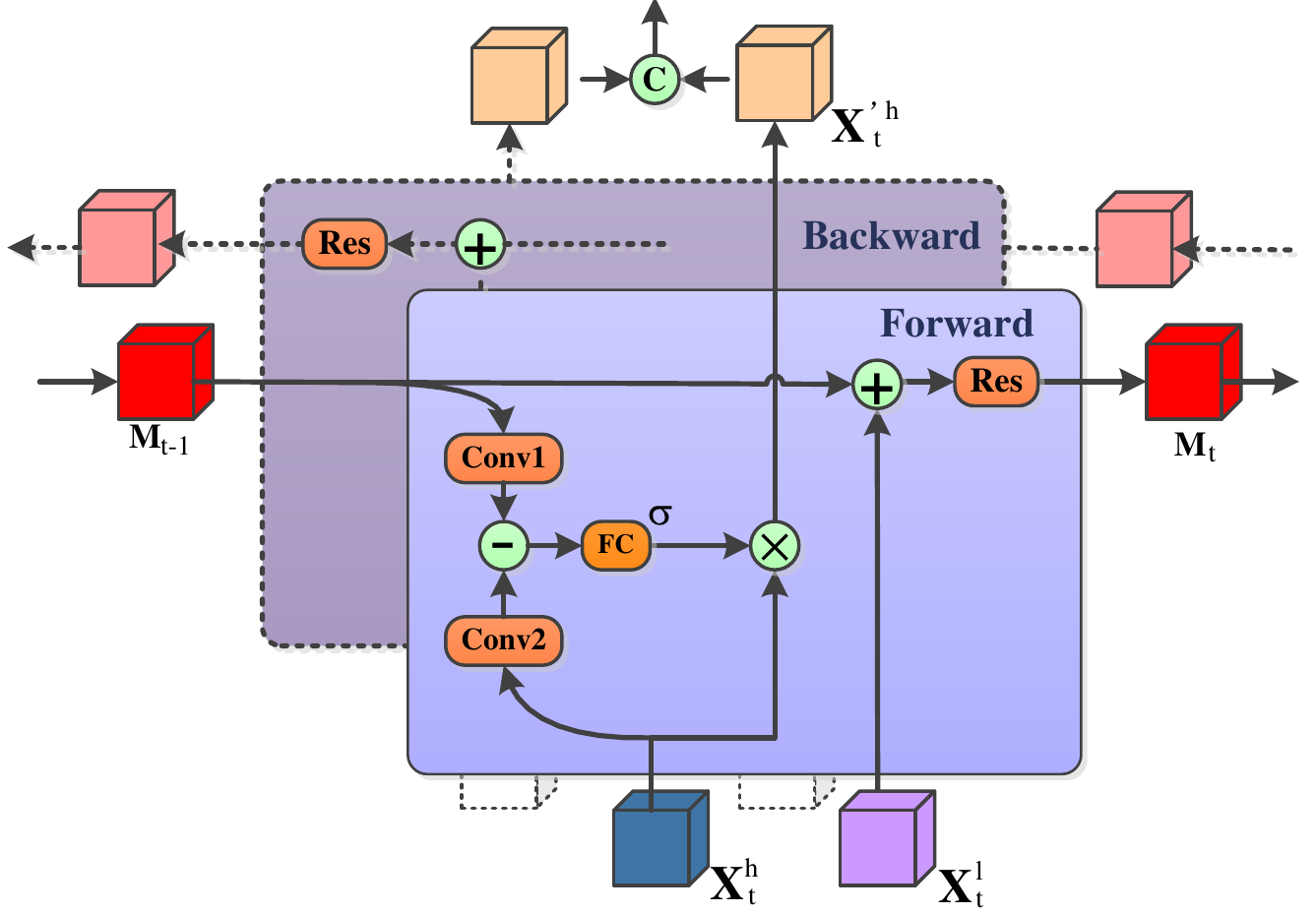} \\
\end{tabular}
}
\vspace{-4mm}
\caption{Enhancement and Memory Unit.}
\label{fig:MemoryUnit}
\vspace{-6mm}
\end{figure}
%-------------------------------------------------------------
\subsection{Temporal Reciprocal Learning}
The temporal feature aggregation plays an important role in video-based person Re-ID.
The GCE can highlight the informative regions in a global view.
However, discontinuous fine-grained cues are easily missed out due to the visual varieties in a long sequence.
To address this problem, we propose a novel Temporal Reciprocal Learning (TRL) mechanism to fully explore the discriminative information from the disentangled high- and low-correlation features.
Considering the frame orders in videos, our TRL is designed for both forward and backward directions.
More specifically, we introduce Enhancement and Memory Units (EMUs) to enhance high-correlation features and accumulate low-correlation features.
Finally, the features passed through the forward and backward directions are integrated as the outputs of our TRL.

\textbf{Enhancement and Memory Unit.}
%The proposed temporal reciprocal learning module includes several Temporal Memory Units (TMU), which are connected in series among the forward and backward process.
%
%The temporal memory unit can sequentially enhance the high-correlation semantic information and accumulate the low-correlation sub-critical cues.
%
%The EMU for time step $t$ of TRL is shown in Fig.~\ref{fig:MemoryUnit}.
%
As illustrated in Fig.~\ref{fig:MemoryUnit}, at the time step $t$, the EMU takes three inputs: the high-correlation features $\textbf{X}_t^{h}$ and the low-correlation features $\textbf{X}_t^{l}$, and the accumulated features $\textbf{M}_{t-1}$ from previous time steps.
%
%In each EMU, there are an enhancer and a memory, which adopt different strategies for high-correlation features and low-correlation features, respectively.
%
%Under the global guidance, the high-correlation features usually highlight the conspicuous and consecutive semantic regions.
%
%For the explore more fine-grained and sub-critical cues, we expect that the accumulative information from low-correlation features could focus on distinct regions and capture the complementary information.
%
%Based on this fact, we consider the accumulative representations $\textbf{M}^{t-1}$ of last frame as a complementary reference.
%
In the enhancement block, we perform subtraction between the high-correlation features $\textbf{X}_t^{h}$ and the accumulated features $\textbf{M}_{t-1}$ to model the difference in semantics.
%
%Thus, the repeated semantic-aware feature between the high-correlation feature $\textbf{X}_t^{h}$ and the accumulative representation $\textbf{M}_{t-1}$, could be suppressed.
%
%In this way, the enhancement block could sequentially determine which channels to focus on in the high-correlation features and accumulative features from low-correlation features.
%
%With such consideration, we design a analogical squeeze and excitation block to reinforce the features on channels in enhancement block.
%
Mathematically, the difference operation is defined as
\begin{equation}\label{se block}
\textbf{D}_t = (f_2(\textbf{M}_{t-1}) - f_1(\textbf{X}_t^{h}))^{2}
\end{equation}
where $f_1$ and $f_2$ represent two individual $1\times 1$ convolution operations with ReLU activation, respectively.
%
%Then, we can obtain the difference of channel statistics $\textbf{D}^t \in R^{C\times H\times W}$ on each pixel-level between $\textbf{M}^{t-1}$ and $\textbf{X}^{h,t}$ at t-th time step.
%
Then, the difference maps $\textbf{D}_t$ are aggregated by GAP to generate an overall response for each channel, \emph{i.e.}, $\textbf{d}_t \in R^C$.
%
%Then two fully connected neural layers following a sigmoid activation function are applied to get the channel attention $\textbf{a}^t\in R^C $
%
We introduce the channel attention for the feature selection, as
\vspace{-2mm}
\begin{equation}\label{channel attention}
\textbf{a}_t = \sigma{(\textbf{W}_c(\textbf{d}_t))}
\end{equation}
\vspace{-4mm}
\begin{equation}\label{channel application}
\textbf{X}_t^{'h} = (1+\textbf{a}_t) \odot \textbf{X}_t^{h} %+ \textbf{X}_t^{h}
\end{equation}
%
%
%\begin{equation}\label{channel application}
%\textbf{M}_{t-1}^{'} = (1+\textbf{a}_t) \odot \textbf{M}_{t-1} %+ \textbf{M}_{t-1}
%\end{equation}
where $\textbf{W}_c$ are the parameters for generating the channel weights.
%
%$\textbf{M}_{t-1}^{'}$ is the final enhancement features.
%
%The averaged channel importance vector is then multiplied with spatially-attended features to obtain the importance-weighted channel activation.
%
%Although the low-correlation features are obtained by inverse operation to high-correlation maps, there are still some meaningful details in them.
%
To fully exploit low-correlation features, we design a memory block to accumulate the fine-grained cues.
Specifically, we first add low-correlation features $\textbf{X}_t^{l}$ at $t$-th frame to the accumulated features $\textbf{M}_{t-1}$ at $t-1$ step.
Then, a residual block~\cite{he2016deep} is utilized for the next EMU.
\begin{equation}\label{veri loss}
\textbf{M}_{t} = Res(\textbf{M}_{t-1} + \textbf{X}_t^{l})
\end{equation}
where $Res$ is the residual block in~\cite{he2016deep}.
In the first time step, $\textbf{M}_0$ is initialized with the mean of $\{\textbf{X}_t^{l}\}_{t=1}^T$.

\textbf{Bi-directional Information Integration.}
In TRL, we design a bi-directional learning mechanism for assembling more robust representations.
The forward and backward directions are relative.
For better understanding, we define the forward direction as the arranged order of video frames.
The backward direction is opposite to forward direction.
%
%In the forward and backward process, the EMUs have the same structure but non-share parameters.
%
With the outputs of EMUs in forward and backward directions, we integrate them as the final video-level representation.
Specifically, the enhanced features ${\textbf{F}}_{t}^{h, 1}$, ${\textbf{F}}_{t}^{h, 2}$, and the accumulated features $\textbf{M}_T^1$  $\textbf{M}_T^2$ in forward and backward are concatenated after GAP.
Then, a fully connected layer is utilized to integrate the concatenated robust representations,
\begin{equation}
\textbf{F}_t^{h} = \textbf{W}_h([\textbf{F}_t^{h,1}, \textbf{F}_t^{h,2}])
\end{equation}
\begin{equation}
\textbf{F}_T^{l} = \textbf{W}_l([\textbf{M}_T^1,\textbf{M}_T^2]).
\end{equation}
With the proposed temporal reciprocal learning mechanism, our method is able to progressively enhance the conspicuous features from high-correlation regions and adaptively mine the sub-critical details from low-correlation regions.
\subsection{Training Schemes}
In our work, we adopt a binary cross entropy loss and the Online Instance Matching loss (OIM)~\cite{xiao2017joint} to train the whole network following~\cite{chen2018video}.
For each probe-gallery video vector pair $\{\textbf{p}_j, \textbf{g}_k\}$ in the training mini-batch, a binary cross entropy loss function can be utilized as
\begin{equation}\label{veri loss}
\!\mathcal{L}\!_{veri}\!=\!-\frac{1}{J}\!\sum_{n=1}^J\!y_jlog([\textbf{p}_j,\textbf{g}_k])\!+\!(1\!-\!y_j\!)log(\!1\!-\![\textbf{p}_j,\textbf{g}_k])
\end{equation}
where $J$ is the number of sampled sequence pairs, $[\cdot,\cdot]$ denotes the similarity estimation function and $[\textbf{p}_j, \textbf{g}_k]\in(0,1)$.
$y_j$ denotes the ground-truth label of $\textbf{p}_j$ and $\textbf{g}_k$.
Note that $y_j=1$ if sequence $\textbf{p}_j$ and $\textbf{g}_k$ belong to the same person, otherwise $y_j=0$.

Meanwhile, in our work, we use a multi-level training objective to deeply supervise our proposed modules, which consists of the frame-level OIM loss and video-level OIM loss.
Instead of the conventional cross-entropy with a multi-class softmax layer, the OIM loss function uses a lookup table to store features of all identifies in the training set.
To learn informative and continuous features from different frames, in the temporal reciprocal learning, the features $\{\textbf{F}_t^{h}\}_{t=1}^{T}$ enhanced by the enhancement block at $t$-th time step, are supervised by a frame-level OIM loss.
% to learn informative and continuous feature from different frames.
%
The frame-level OIM loss can be defined as:
\begin{equation}\label{frame-wise oim loss}
\!\mathcal{L}_f\!=\!-\frac{1}{N\!\times\! T}\!\sum_{n=1}^N\!\sum_{t=1}^T\!\sum_{i=1}^I\!y^{i}_{t,n}log(\frac{e^{\textbf{W}_i\textbf{F}_{t, n}^{h}}}{\!\sum_{j=1}^I\!e^{\textbf{W}_j\textbf{F}_{t,n}^{h}}})
\end{equation}
where $\textbf{F}_{t,n}^{h}$ indicates the enhanced high-correlation feature vector of the $t$-th image in $n$-th video.
If the $t$-th image in $n$-th video belongs to the $i$-th person, $y^{i}_{t,n} = 1$, otherwise $y^{i}_{t,n} = 0$.
$\textbf{W}_i$ are the coefficients associated with the feature embedding of the $i$-th person, which are online updated with the frame-wise feature vectors of the $i$-th person.
%
%They are obtained by using an online updated buffer and measuring similarities between the current person and all the other persons in the feature buffer with inner product.
%
Meanwhile, the feature $\textbf{F}_T^{l}$ accumulated by the memory block at last time step, is supervised by video-level OIM loss, which attempts to progressively collect all the sub-critical details from the low-correlation regions.
\begin{equation}\label{video-wise oim loss}
\mathcal{L}_v = - \frac{1}{N} \sum_{n=1}^N \sum_{i=1}^I y^{i}_nlog(\frac{e^{\textbf{W}_i\textbf{F}_T^{l}}}{\sum_{j=1}^I e^{\textbf{W}_j\textbf{F}_T^{l}}})
\end{equation}

The total loss is a combination of the frame-level OIM loss, the video-level OIM loss and the verification loss.
\begin{equation}\label{total loss}
\mathcal{L} = \lambda_1\mathcal{L}_f + \lambda_2\mathcal{L}_v + \lambda_3\mathcal{L}_{veri}
\end{equation}
\section{Experiments}
\subsection{Datasets and Evaluation Protocols}
To evaluate the performance of our proposed method, we adopt three widely-used benchmarks, \emph{i.e.}, iLIDS-VID~\cite{wang2014person}, PRID-2011~\cite{hirzer2011person} and MARS~\cite{zheng2016mars}.
iLIDS-VID~\cite{wang2014person} dataset is a small dataset, which consists of 600 video sequences of 300 different identities.
Two cameras are used to collect images.
Each video sequence contains 23 to 192 frames.
PRID-2011~\cite{hirzer2011person} dataset consists of 400 image sequences for 200 identities from two non-overlapping cameras.
The sequence lengths range from 5 to 675 frames, with an average of 100.
Following previous practice~\cite{wang2014person}, we only utilize the sequence pairs with more than 21 frames.
MARS~\cite{zheng2016mars} is one of large-scale datasets, and consists of 1,261 identities around 18,000 video sequences.
All the video sequences are captured by at least 2 cameras.
%
%In order to simulate to actual detect conditions, there are round 3,200 distractors sequences in the dataset.
%

For evaluation, we follow previous works and adopt the Cumulative Matching Characteristic (CMC) table and mean Average Precision (mAP) to evaluate the performance. %for all the datasets.
%For more details, we refer readers to the original paper~\cite{zheng2016person}.
%For ease of comparison, we only report the cumulative re-identification accuracy at selected ranks.
In terms of iLIDS-VID and PRID2011, we only report the cumulative re-identification accuracy because that there only contains a single correct match in the gallery set.
\begin{table*}
\begin{center}
\doublerulesep=0.5pt
%\vspace{-2mm}
\caption{Comparison with state-of-the-art video-based person Re-ID methods on MARS, iLIDS-VID and PRID2011.}
\label{table:soa method}
%\vspace{-2mm}
\resizebox{1\textwidth}{!}
{
\begin{tabular}{|c|c|c|c|c|c|c|c|c|c|c|c|c|c|c|c|c|c|c|c|c|c|c|c|c|c|c|c|c|c|c|c|c|c|c|c|c|c|c|c|c|c|c|c|c|c|c|c|c|c|c|c|c|c|c|c|c|c|c|c|c|c|c|c|c|c|c|c|c|c|c|c|c|c|c|c|c|c|c|c|c|c|c|c}
\hline
\multicolumn{8}{c}{}
&\multicolumn{8}{c|}{}
&\multicolumn{16}{c|}{MARS}
&\multicolumn{12}{c|}{iLIDS-VID}
&\multicolumn{12}{c}{PRID2011}
\\
\multicolumn{8}{c|}{Methods}
&\multicolumn{8}{c}{Source}
&\multicolumn{4}{|c}{mAP}
&\multicolumn{4}{c}{Rank-1}
&\multicolumn{4}{c}{Rank-5}
&\multicolumn{4}{c|}{Rank-20}
&\multicolumn{4}{c}{Rank-1}
&\multicolumn{4}{c}{Rank-5}
&\multicolumn{4}{c|}{Rank-20}
&\multicolumn{4}{c}{Rank-1}
&\multicolumn{4}{c}{Rank-5}
&\multicolumn{4}{c}{Rank-20}
\\
\hline
\multicolumn{8}{l|}{SeeForest~\cite{zhou2017see}}
&\multicolumn{8}{c|}{CVPR17}
&\multicolumn{4}{c}{50.7}
&\multicolumn{4}{c}{70.6}
&\multicolumn{4}{c}{90.0}
&\multicolumn{4}{c}{97.6}
&\multicolumn{4}{|c}{55.2}
&\multicolumn{4}{c}{86.5}
&\multicolumn{4}{c|}{97.0}
&\multicolumn{4}{c}{79.4}
&\multicolumn{4}{c}{94.4}
&\multicolumn{4}{c}{99.3}
\\
\multicolumn{8}{l|}{ASTPN~\cite{xu2017jointly}}
&\multicolumn{8}{c|}{ICCV17}
&\multicolumn{4}{c}{-}
&\multicolumn{4}{c}{44}
&\multicolumn{4}{c}{70}
&\multicolumn{4}{c}{81}
&\multicolumn{4}{|c}{62}
&\multicolumn{4}{c}{86}
&\multicolumn{4}{c|}{98}
&\multicolumn{4}{c}{77}
&\multicolumn{4}{c}{95}
&\multicolumn{4}{c}{99}
\\
\multicolumn{8}{l|}{Snippet~\cite{chen2018video}}
&\multicolumn{8}{c|}{CVPR18}
&\multicolumn{4}{c}{76.1}
&\multicolumn{4}{c}{86.3}
&\multicolumn{4}{c}{94.7}
&\multicolumn{4}{c}{98.2}
&\multicolumn{4}{|c}{85.4}
&\multicolumn{4}{c}{96.7}
&\multicolumn{4}{c|}{99.5}
&\multicolumn{4}{c}{93.0}
&\multicolumn{4}{c}{99.3}
&\multicolumn{4}{c}{100}
\\
\multicolumn{8}{l|}{STAN~\cite{li2018diversity}}
&\multicolumn{8}{c|}{CVPR18}
&\multicolumn{4}{c}{65.8}
&\multicolumn{4}{c}{82.3}
&\multicolumn{4}{c}{-}
&\multicolumn{4}{c}{-}
&\multicolumn{4}{|c}{80.2}
&\multicolumn{4}{c}{-}
&\multicolumn{4}{c|}{-}
&\multicolumn{4}{c}{93.2}
&\multicolumn{4}{c}{-}
&\multicolumn{4}{c}{-}
\\
\multicolumn{8}{l|}{STMP~\cite{liu2019spatial}}
&\multicolumn{8}{c|}{AAAI19}
&\multicolumn{4}{c}{72.7}
&\multicolumn{4}{c}{84.4}
&\multicolumn{4}{c}{93.2}
&\multicolumn{4}{c}{96.3}
&\multicolumn{4}{|c}{84.3}
&\multicolumn{4}{c}{96.8}
&\multicolumn{4}{c|}{99.5}
&\multicolumn{4}{c}{92.7}
&\multicolumn{4}{c}{98.8}
&\multicolumn{4}{c}{99.8}
\\
\multicolumn{8}{l|}{M3D~\cite{li2019multi}}
&\multicolumn{8}{c|}{AAAI19}
&\multicolumn{4}{c}{74.0}
&\multicolumn{4}{c}{84.3}
&\multicolumn{4}{c}{93.8}
&\multicolumn{4}{c}{97.7}
&\multicolumn{4}{|c}{74.0}
&\multicolumn{4}{c}{94.3}
&\multicolumn{4}{c|}{-}
&\multicolumn{4}{c}{94.4}
&\multicolumn{4}{c}{100}
&\multicolumn{4}{c}{-}
\\
\multicolumn{8}{l|}{STA~\cite{fu2019sta}}
&\multicolumn{8}{c|}{AAAI19}
&\multicolumn{4}{c}{80.8}
&\multicolumn{4}{c}{86.3}
&\multicolumn{4}{c}{95.7}
&\multicolumn{4}{c}{98.1}
&\multicolumn{4}{|c}{-}
&\multicolumn{4}{c}{-}
&\multicolumn{4}{c|}{-}
&\multicolumn{4}{c}{-}
&\multicolumn{4}{c}{-}
&\multicolumn{4}{c}{-}
\\
\multicolumn{8}{l|}{Attribute~\cite{zhao2019attribute}}
&\multicolumn{8}{c|}{CVPR19}
&\multicolumn{4}{c}{78.2}
&\multicolumn{4}{c}{87.0}
&\multicolumn{4}{c}{95.4}
&\multicolumn{4}{c}{\textbf{98.7}}
&\multicolumn{4}{|c}{86.3}
&\multicolumn{4}{c}{87.4}
&\multicolumn{4}{c|}{99.7}
&\multicolumn{4}{c}{93.9}
&\multicolumn{4}{c}{99.5}
&\multicolumn{4}{c}{100}
\\
\multicolumn{8}{l|}{VRSTC~\cite{hou2019vrstc}}
&\multicolumn{8}{c|}{CVPR19}
&\multicolumn{4}{c}{82.3}
&\multicolumn{4}{c}{88.5}
&\multicolumn{4}{c}{96.5}
&\multicolumn{4}{c}{97.4}
&\multicolumn{4}{|c}{83.4}
&\multicolumn{4}{c}{95.5}
&\multicolumn{4}{c|}{99.5}
&\multicolumn{4}{c}{-}
&\multicolumn{4}{c}{-}
&\multicolumn{4}{c}{-}
\\
\multicolumn{8}{l|}{GLTR~\cite{li2019global}}
&\multicolumn{8}{c|}{ICCV19}
&\multicolumn{4}{c}{78.5}
&\multicolumn{4}{c}{87.0}
&\multicolumn{4}{c}{95.8}
&\multicolumn{4}{c}{98.2}
&\multicolumn{4}{|c}{86.0}
&\multicolumn{4}{c}{98.0}
&\multicolumn{4}{c|}{-}
&\multicolumn{4}{c}{95.5}
&\multicolumn{4}{c}{\textbf{100}}
&\multicolumn{4}{c}{-}
\\
\multicolumn{8}{l|}{COSAM~\cite{subramaniam2019co}}
&\multicolumn{8}{c|}{ICCV19}
&\multicolumn{4}{c}{79.9}
&\multicolumn{4}{c}{84.9}
&\multicolumn{4}{c}{95.5}
&\multicolumn{4}{c}{97.9}
&\multicolumn{4}{|c}{79.6}
&\multicolumn{4}{c}{95.3}
&\multicolumn{4}{c|}{-}
&\multicolumn{4}{c}{-}
&\multicolumn{4}{c}{-}
&\multicolumn{4}{c}{-}
\\
\multicolumn{8}{l|}{MGRA~\cite{zhang2020multi}}
&\multicolumn{8}{c|}{CVPR20}
&\multicolumn{4}{c}{\textbf{85.9}}
&\multicolumn{4}{c}{88.8}
&\multicolumn{4}{c}{\textbf{97.0}}
&\multicolumn{4}{c}{\underline{98.5}}
&\multicolumn{4}{|c}{\underline{88.6}}
&\multicolumn{4}{c}{\underline{98.0}}
&\multicolumn{4}{c|}{\underline{99.7}}
&\multicolumn{4}{c}{\underline{95.9}}
&\multicolumn{4}{c}{99.7}
&\multicolumn{4}{c}{\underline{100}}
\\
\multicolumn{8}{l|}{STGCN~\cite{yang2020spatial}}
&\multicolumn{8}{c|}{CVPR20}
&\multicolumn{4}{c}{83.7}
&\multicolumn{4}{c}{89.9}
&\multicolumn{4}{c}{-}
&\multicolumn{4}{c}{-}
&\multicolumn{4}{|c}{-}
&\multicolumn{4}{c}{-}
&\multicolumn{4}{c|}{-}
&\multicolumn{4}{c}{-}
&\multicolumn{4}{c}{-}
&\multicolumn{4}{c}{-}
\\
\multicolumn{8}{l|}{AFA~\cite{chen2020temporal}}
&\multicolumn{8}{c|}{ECCV20}
&\multicolumn{4}{c}{82.9}
&\multicolumn{4}{c}{\underline{90.2}}
&\multicolumn{4}{c}{96.6}
&\multicolumn{4}{c}{-}
&\multicolumn{4}{|c}{88.5}
&\multicolumn{4}{c}{96.8}
&\multicolumn{4}{c|}{99.7}
&\multicolumn{4}{c}{-}
&\multicolumn{4}{c}{-}
&\multicolumn{4}{c}{-}
\\
\multicolumn{8}{l|}{TCLNet~\cite{hou2020temporal}}
&\multicolumn{8}{c|}{ECCV20}
&\multicolumn{4}{c}{\underline{85.1}}
&\multicolumn{4}{c}{89.8}
&\multicolumn{4}{c}{-}
&\multicolumn{4}{c}{-}
&\multicolumn{4}{|c}{86.6}
&\multicolumn{4}{c}{-}
&\multicolumn{4}{c|}{-}
&\multicolumn{4}{c}{-}
&\multicolumn{4}{c}{-}
&\multicolumn{4}{c}{-}
\\
\hline
\multicolumn{8}{l|}{Ours}
&\multicolumn{8}{c|}{--}
&\multicolumn{4}{c}{84.8}
&\multicolumn{4}{c}{\textbf{91.0}}
&\multicolumn{4}{c}{\underline{96.7}}
&\multicolumn{4}{c}{98.4}
&\multicolumn{4}{|c}{\textbf{90.4}}
&\multicolumn{4}{c}{\textbf{98.3}}
&\multicolumn{4}{c|}{\textbf{99.8}}
&\multicolumn{4}{c}{\textbf{96.2}}
&\multicolumn{4}{c}{\underline{99.7}}
&\multicolumn{4}{c}{\textbf{100}}
\\
\hline
\end{tabular}
}
\vspace{-4mm}
\end{center}
\end{table*}
\subsection{Implementation Details}
We implement our framework based on the Pytorch\footnote{https://pytorch.org/} toolbox.
The experimental devices include an Intel i4790 CPU and two NVIDIA GTX 2080ti GPUs (12G memory).
To generate training sequences, we employ the RRS strategy~\cite{fu2019sta}, and divide each video sequence into 8 chunks with equal duration.
Experimentally, we set the batchsize = 16 and $T$ = 8.
Each image in a sequence is resized to 256$\times$128 and the input sequences are augmented by random cropping, horizontal flipping and random erasing.
To provide a number of positive and negative sequence pairs in each training mini-batch, we randomly sampled the half batchsize sequences firstly.
For one sampled sequences, we select another sequence with the same identify but under different cameras to fill the total batch.
In this way, there is at least one positive sample for any sequence in a mini-batch.
The ResNet-50~\cite{he2016deep} pre-trained on the ImageNet dataset~\cite{deng2009imagenet} is used as our backbone network.
Following previous works~\cite{sun2018beyond}, we remove the last spatial down-sampling operation to increase the feature resolution.
During training, we train our network for 50 epochs combining with the multi-level OIM losses and a binary cross-entropy loss.
The whole network is updated by stochastic gradient descent~\cite{bottou2010large} algorithm with an initial learning rate of $10^{-3}$, weight decay of $5\times10^{-4}$ and nesterov momentum of 0.9.
The learning rate is decayed by 10 at every 15 epochs.
%-------------------------------------------------------------
\subsection{Comparison with State-of-the-arts}
In this section, we compare the proposed approach with other state-of-the-art methods on three video-based person Re-ID benchmarks.
Experimental results are reported in Tab.~\ref{table:soa method}.
On MARS dataset, the mAP and Rank-1 accuracy of our proposed method are $84.8\%$ and $91.0\%$, respectively.
Besides, our method achieves $90.4\%$ and $96.2\%$ of the Rank-1 accuracy on iLIDS-VID dataset and PRID2011 dataset.
The Rank-1 accuracy of our method outperforms all the compared methods, showing significant improvements over several state-of-the-art methods.
We note that the MGRA~\cite{zhang2020multi} also employs the global view for the video-based person Re-ID task.
It gains remarkable $85.9\%$ mAP on MARS dataset.
Different from MGRA, our method utilizes the global representations to estimate two correlation maps for the feature disentanglement on spatial features.
%
%Under the guidance of a global view, the correlation maps are estimated with minimal parameters and computational costs than MGRA.
%
With the reciprocal learning, our method could fully take advantages of the disentangled features, and explore more informative and fine-grained cues via the high-correlation maps and low-correlation maps.
Thereby, our method surpasses MGRA by $2.2\%$, $1.8\%$ and $0.3$ in terms of Rank-1 accuracy on MARS, iLIDS-VID and PRID2011, respectively.
Meanwhile, it is worth noting that those methods, ASTPN~\cite{mclaughlin2016recurrent}, STMP~\cite{liu2019spatial} and GLTR~\cite{li2019global}, explore the temporal learning for video-based person Re-ID.
ASTPN~\cite{mclaughlin2016recurrent} utilizes a temporal RNN to model the temporal information for video representations.
STMP~\cite{liu2019spatial} introduces a refining recurrent unit to recover the missing parts by referring historical frames.
GLTR~\cite{li2019global} employs dilated temporal convolutions to capture the multi-granular temporal dependencies and aggregates short and long-term temporal cues for global-local temporal representations.
Compared with these methods, our proposed method achieves better results on three public datasets.
More specifically, compared with GLTR~\cite{li2019global}, our method improves the performances by $6.3\%$ and $4.0\%$ in terms of mAP and Rank-1 accuracy on MARS dataset.
In summary, compared with existing methods, our method utilizes the global information to guide the feature disentanglement.
In additional, we adopt two strategies to mine richer cues for temporal learning, which can fully exploit the spatial-temporal information for more discriminative video representations.
These experimental results validate the superiority of our method.
%------------------------------------------------
\subsection{Ablation Study}
In this subsection, we conduct experiments to verify the effectiveness of the proposed methods.
All the models are trained and evaluated on MARS, iLIDS-VID and PRID2011 datasets.
Results are shown in Tab.~\textcolor{red}{1-5}.
In these tables, ``Baseline'' represents the backbone trained only with video-level OIM loss on the global branch, in which TAP and GAP are applied on the frame-level features.

\textbf{Effects of Key Components.}
The ablation results of key components are reported in Tab.~\ref{table:ablation of main}.
In this table, $F^g$ denotes the global feature vector without disentanglements.
$F^l$ denotes the final feature vector with disentangled low-correlation features, and is supervised with a video-level OIM loss.
$F^h$ denotes the final feature vector with disentangled high-correlation features, and is supervised with a frame-level OIM loss.
``+GCE'' means that we add the global-guided correlation estimation to guide the disentanglement of spatial features.
One can see that the performance has a significant improvement after disentanglement.
The disentangled high-correlation features increase the Rank-1 accuracy by $1.6\%$, $2.4\%$ and $1.5\%$ on MARS, iLIDS-VID and PRID2011, respectively.
Thus, it is beneficial to guide the feature disentanglement under a global view.
%-------------------------------------------------------------
%Considering that the low-correlated regions in a frame could be meaningful for identification, we design a novel temporal reciprocal learning mechanism to execute different learning strategies for disentangled features.
%
``+TRL'' means that the temporal reciprocal learning with bi-directions is used to enhance and accumulate temporal information.
Compared with the ``+GCE'' model, our proposed TRL mechanism can further improve the mAP by $1.5\%$ and the Rank-1 accuracy by $0.9\%$ on MARS.
As shown in Tab.~\ref{table:ablation of main}, the combination of the low- and high-correlation features can further boost the performance.
%
%Heavily, the increments benefit from jointly exploiting the conspicuous and informative characteristics and the sub-critical details.
%
The above results clearly demonstrate the effectiveness of our proposed GCE and TRL modules.
\begin{table}
\caption{Ablation results of key components on three benchmarks.}
\label{table:ablation of main}
\vspace{-4mm}
\begin{center}
\doublerulesep=0.1pt
\resizebox{0.48\textwidth}{!}
{
\begin{tabular}{|c|c|c|c|c|c|c|c|c|c|c|c|c|c|c|c|c|c|c|c|c|c|c|c|c|c|c|c|c|c|c|c|c|c|c|c|c|c|c|c|c|c|c|c|c|c|c|c|c|c|c|c|c|c|c|c|c|c|c|c|c|c|c|c|c|c|c|c|c|c|c|c|c|c|c|}
\hline
\multicolumn{8}{l}{}
&\multicolumn{8}{|l}{}
&\multicolumn{8}{|c|}{MARS}
&\multicolumn{8}{|c}{iLIDS-VID}
&\multicolumn{8}{|c}{PRID2011}
\\
\multicolumn{8}{l}{Methods}
&\multicolumn{8}{|c}{Feat. to test}
&\multicolumn{4}{|c}{mAP}
&\multicolumn{4}{c|}{Rank-1}
&\multicolumn{4}{c}{Rank-1}
&\multicolumn{4}{c|}{Rank-5}
&\multicolumn{4}{c}{Rank-1}
&\multicolumn{4}{c}{Rank-5}
\\
\hline
\hline
\multicolumn{8}{l}{Baseline}
&\multicolumn{8}{|c}{$F^g$}
&\multicolumn{4}{|c}{81.2}
&\multicolumn{4}{c|}{88.5}
&\multicolumn{4}{c}{87.1}
&\multicolumn{4}{c}{97.2}
&\multicolumn{4}{|c}{93.5}
&\multicolumn{4}{c}{98.7}
\\
\hline
\hline
\multicolumn{8}{l}{}
&\multicolumn{8}{|c}{$F^l$}
&\multicolumn{4}{|c}{81.9}&\multicolumn{4}{c}{88.9}
&\multicolumn{4}{|c}{88.3}
&\multicolumn{4}{c}{97.4}
&\multicolumn{4}{|c}{93.9}
&\multicolumn{4}{c}{99.0}
\\
\multicolumn{8}{l}{+ GCE}
&\multicolumn{8}{|c}{$F^h$}
&\multicolumn{4}{|c}{83.0}&\multicolumn{4}{c}{89.5}
&\multicolumn{4}{|c}{89.2}
&\multicolumn{4}{c}{98.0}
&\multicolumn{4}{|c}{94.7}
&\multicolumn{4}{c}{99.2}
\\
\multicolumn{8}{l}{}
&\multicolumn{8}{|c}{$F^l, F^h$}
&\multicolumn{4}{|c}{83.3}&\multicolumn{4}{c}{90.1}
&\multicolumn{4}{|c}{89.5}
&\multicolumn{4}{c}{97.9}
&\multicolumn{4}{|c}{95.0}
&\multicolumn{4}{c}{99.5}
\\
\hline
\hline
\multicolumn{8}{l}{}
&\multicolumn{8}{|c}{$F^l$}
&\multicolumn{4}{|c}{82.2}&\multicolumn{4}{c}{88.4}
&\multicolumn{4}{|c}{88.6}
&\multicolumn{4}{c}{97.8}
&\multicolumn{4}{|c}{94.8}
&\multicolumn{4}{c}{99.6}
\\
\multicolumn{8}{l}{+ TRL}
&\multicolumn{8}{|c}{$F^h$}
&\multicolumn{4}{|c}{84.0}&\multicolumn{4}{c}{90.4}
&\multicolumn{4}{|c}{\textbf{90.5}}
&\multicolumn{4}{c}{97.9}
&\multicolumn{4}{|c}{95.4}
&\multicolumn{4}{c}{99.6}
\\
\multicolumn{8}{l}{}
&\multicolumn{8}{|c}{$F^h, F^l$}
&\multicolumn{4}{|c}{\textbf{84.8}}&\multicolumn{4}{c}{\textbf{91.0}}
&\multicolumn{4}{|c}{90.4}
&\multicolumn{4}{c}{\textbf{98.3}}
&\multicolumn{4}{|c}{\textbf{96.2}}
&\multicolumn{4}{c}{\textbf{99.7}}
\\
\hline
\hline
\end{tabular}
}
\vspace{-4mm}
\end{center}
\end{table}
%------------------------------------------------------
\begin{table}
\caption{Ablation results of EMU on MARS.}
\label{table: ablation of EMU}
\vspace{-2mm}
\begin{center}
\doublerulesep=0.1pt
\resizebox{0.48\textwidth}{!}
{
\begin{tabular}{|c|c|c|c|c|c|c|c|c|c|c|c|c|c|c|c|c|c|c|c|c|c|c|c|c|c|c|c|c|c|c|c|c|c|c|c|c|c|c|c|}
\hline
\multicolumn{8}{l|}{Methods}
&\multicolumn{4}{c}{mAP}&\multicolumn{4}{c}{Rank-1}
&\multicolumn{4}{c}{Rank-5}&\multicolumn{4}{c}{Rank-20}
\\
\hline
\hline
\multicolumn{8}{l|}{Baseline}
&\multicolumn{4}{c}{81.2}&\multicolumn{4}{c}{88.5}
&\multicolumn{4}{c}{95.5}&\multicolumn{4}{c}{97.9}
\\
\hline
\hline
\multicolumn{8}{l|}{GRL}
&\multicolumn{4}{c}{84.8}&\multicolumn{4}{c}{91.0}
&\multicolumn{4}{c}{96.7}&\multicolumn{4}{c}{98.4}
\\
\multicolumn{8}{l|}{- Memory Block}
&\multicolumn{4}{c}{84.2}&\multicolumn{4}{c}{90.2}
&\multicolumn{4}{c}{96.3}&\multicolumn{4}{c}{98.2}
\\
\multicolumn{8}{l|}{- Enhancement Block}
&\multicolumn{4}{c}{83.4}&\multicolumn{4}{c}{90.1}
&\multicolumn{4}{c}{96.5}&\multicolumn{4}{c}{98.3}
\\
\hline
\hline
\end{tabular}
}
\vspace{-4mm}
\end{center}
\end{table}
%-------------------------------------------------------------------------------------
\begin{table}
\caption{Ablation results of the sequence length.}
\label{table: ablation of the length}
\vspace{-2mm}
\begin{center}
\doublerulesep=0.1pt
\resizebox{0.48\textwidth}{!}
{
\begin{tabular}{|c|c|c|c|c|c|c|c|c|c|c|c|c|c|c|c|c|c|c|c|c|c|c|c|c|c|c|c|c|c|c|c|c|c|c|c|c|c|c|c|}
\hline
\multicolumn{8}{l|}{Methods}
&\multicolumn{4}{l|}{Length}
&\multicolumn{4}{c}{mAP}&\multicolumn{4}{c}{Rank-1}
&\multicolumn{4}{c}{Rank-5}&\multicolumn{4}{c}{Rank-20}
\\
\hline
\hline
\multicolumn{8}{l|}{Baseline}
&\multicolumn{4}{c|}{8}
&\multicolumn{4}{c}{81.2}&\multicolumn{4}{c}{88.5}
&\multicolumn{4}{c}{95.5}&\multicolumn{4}{c}{97.9}
\\
\hline
\hline
\multicolumn{8}{l|}{GRL}
&\multicolumn{4}{c|}{4}
&\multicolumn{4}{c}{83.0}&\multicolumn{4}{c}{89.4}
&\multicolumn{4}{c}{96.1}&\multicolumn{4}{c}{98.2}
\\
\multicolumn{8}{c|}{}
&\multicolumn{4}{c|}{6}
&\multicolumn{4}{c}{83.5}&\multicolumn{4}{c}{90.1}
&\multicolumn{4}{c}{96.7}&\multicolumn{4}{c}{98.4}
\\
\multicolumn{8}{l|}{}
&\multicolumn{4}{c|}{8}
&\multicolumn{4}{c}{84.8}&\multicolumn{4}{c}{91.0}
&\multicolumn{4}{c}{96.7}&\multicolumn{4}{c}{98.4}
\\
\multicolumn{8}{l|}{}
&\multicolumn{4}{c|}{10}
&\multicolumn{4}{c}{83.9}&\multicolumn{4}{c}{90.2}
&\multicolumn{4}{c}{96.5}&\multicolumn{4}{c}{98.4}
\\
\hline
\hline
\end{tabular}
}
\vspace{-4mm}
\end{center}
\end{table}
%---------------------------------------------------
\begin{table}
\caption{Ablation results of the direction order.}
\label{table: ablation of the direction}
\vspace{-2mm}
\begin{center}
\doublerulesep=0.1pt
\resizebox{0.48\textwidth}{!}
{
\begin{tabular}{|c|c|c|c|c|c|c|c|c|c|c|c|c|c|c|c|c|c|c|c|c|c|c|c|c|c|c|c|c|c|c|c|c|c|c|c|c|c|c|c|}
\hline
\multicolumn{8}{l|}{Methods}
&\multicolumn{4}{l|}{Direction}
&\multicolumn{4}{c}{mAP}&\multicolumn{4}{c}{Rank-1}
&\multicolumn{4}{c}{Rank-5}&\multicolumn{4}{c}{Rank-20}
\\
\hline
\hline
\multicolumn{8}{l|}{Baseline}
&\multicolumn{4}{l|}{}
&\multicolumn{4}{c}{81.2}&\multicolumn{4}{c}{88.5}
&\multicolumn{4}{c}{95.5}&\multicolumn{4}{c}{97.9}
\\
\hline
\hline
\multicolumn{8}{l|}{GRL}
&\multicolumn{4}{l|}{Forward}
&\multicolumn{4}{c}{84.0}&\multicolumn{4}{c}{89.9}
&\multicolumn{4}{c}{96.5}&\multicolumn{4}{c}{98.3}
\\
\multicolumn{8}{c|}{}
&\multicolumn{4}{l|}{Backward}
&\multicolumn{4}{c}{83.7}&\multicolumn{4}{c}{90.0}
&\multicolumn{4}{c}{96.5}&\multicolumn{4}{c}{98.3}
\\
\multicolumn{8}{l|}{}
&\multicolumn{4}{c|}{Bi-direction}
&\multicolumn{4}{c}{84.8}&\multicolumn{4}{c}{91.0}
&\multicolumn{4}{c}{96.7}&\multicolumn{4}{c}{98.4}
\\
\hline
\hline
\end{tabular}
}
\vspace{-6mm}
\end{center}
\end{table}
%-------------------------------------------------------------------------------------

%------------------------------------------------------
\par\textbf{Effects of Enhancement and Memory Unit.}
We also perform experiments to verify the effectiveness of EMU.
The results on MARS are shown in Tab.~\ref{table: ablation of EMU}.
‘‘GRL'' means that our proposed GRL approach with complete EMUs.
‘‘- Memory Block'' denotes that the memory blocks in EMUs are removed.
The results show that, there are slight decreases in terms of mAP and Rank-1 accuracy on MARS.
‘‘- Enhancement Block'' denotes that the enhancement blocks in EMUs are removed, in which the high-correlation features $\textbf{X}_t^{h}$ are supervised by the frame-level OIM loss without channel attention.
The mAP and Rank-1 accuracy drop with $1.4\%$ and $0.9\%$ on MARS.
From the results, we can find that both our enhancement block and memory block are beneficial to learn more discriminative spatial features.
%------------------------------------------------------

\textbf{Effects of Different Sequence Lengths.}
We train and test our bi-directional global-guided reciprocal learning with various sequence lengths $T$.
The results are shown in Tab.~\ref{table: ablation of the length}.
From the results, we can see that increasing the length of sequence gains better performance and the length of $8$ gets best performance.
One possible reason is that our temporal reciprocating learning could capture more fine-grained cues with the increments of the sequence length.
However, too long sequences are not good for training the temporal reciprocal learning module.
%------------------------------------------------------

\textbf{Effects of Temporal Directions.}
We perform additional experiments to verify the effectiveness of temporal directions in GRL.
As shown in Tab.~\ref{table: ablation of the direction}, the proposed temporal learning with forward or backward direction gains similar results.
Besides, the bi-directional reciprocal learning shows higher performances, which benefits from the combination of forward and backward temporal learning.
The effectiveness indicates that the aggregated features by reciprocating learning are more robust for identification.
%-----------------------------------------
\begin{figure*}
\centering
\resizebox{1\textwidth}{!}
{
\begin{tabular}{@{}c@{}c@{}}
\includegraphics[width=0.9\linewidth,height=0.45\linewidth]{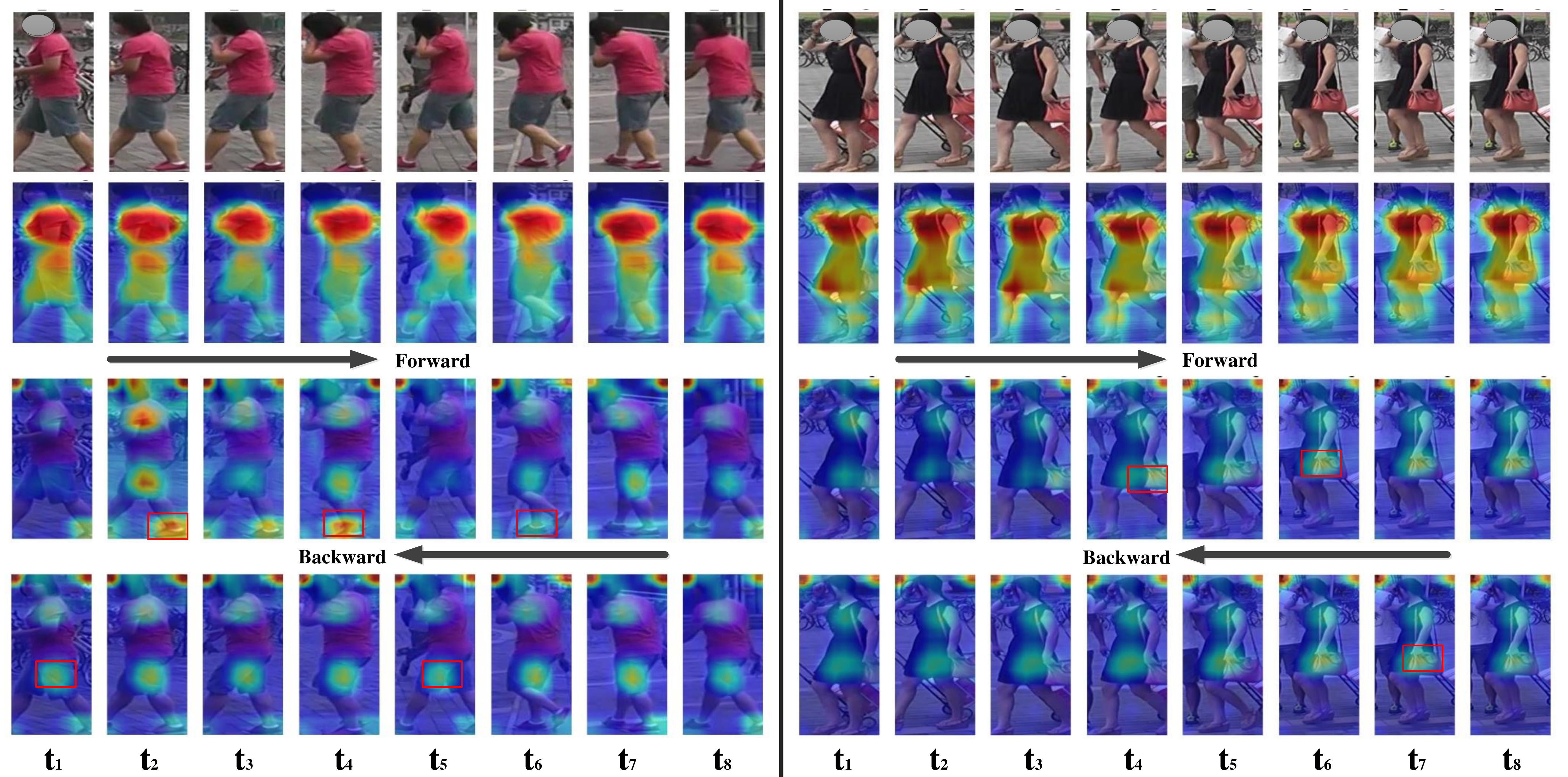} \\
\end{tabular}
}
%\vspace{2mm}
\caption{The visualization of the high-correlation maps and the accumulated low-correlation features at different time steps.
The top images are raw images in video sequences.
The heat maps in the second row are high-correlation maps $\{R_t\}_{t=1}^{T=8}$. %guided by the global view.
The heat maps in the third and fourth rows are channel activation maps of accumulated features, $\{M_t^1\}_{t=1}^{T=8}$ and $\{M_t^2\}_{t=1}^{T=8}$ in the forward and backward process.
}
\label{fig:visualization}
\vspace{-2mm}
\end{figure*}
%--------------------------------
%------------------------------------------------------
\begin{table}
\caption{Ablation results of multi-level OIM losses on MARS.}
\vspace{-4mm}
\label{table: ablation of oim}
\begin{center}
\doublerulesep=0.05pt
\resizebox{0.48\textwidth}{!}
{
\begin{tabular}{|c|c|c|c|c|c|c|c|c|c|c|c|c|c|c|c|c|c|c|c|c|c|c|c|c|c|c|c|c|c|c|c|c|c|c|c|c|c|c|c|}
\hline
\multicolumn{8}{l|}{Methods}
&\multicolumn{4}{c|}{Losses}
&\multicolumn{4}{c}{mAP}&\multicolumn{4}{c}{Rank-1}
&\multicolumn{4}{c}{Rank-5}&\multicolumn{4}{c}{Rank-20}
\\
\hline
\hline
\multicolumn{8}{l|}{}
&\multicolumn{4}{c|}{V-OIM}
&\multicolumn{4}{c}{82.6}&\multicolumn{4}{c}{89.5}
&\multicolumn{4}{c}{96.3}&\multicolumn{4}{c}{98.1}
\\
\multicolumn{8}{l|}{GRL}
&\multicolumn{4}{c|}{F-OIM}
&\multicolumn{4}{c}{83.5}&\multicolumn{4}{c}{90.1}
&\multicolumn{4}{c}{96.4}&\multicolumn{4}{c}{98.3}
\\
\multicolumn{8}{l|}{}
&\multicolumn{4}{c|}{V\&F-OIM}
&\multicolumn{4}{c}{84.8}&\multicolumn{4}{c}{91.0}
&\multicolumn{4}{c}{96.7}&\multicolumn{4}{c}{98.4}
\\
\hline
\hline
\end{tabular}
}
\vspace{-6mm}
\end{center}
\end{table}
%-----------------------------------------------------------------------

\textbf{Effects of Multi-level OIM Losses.}
The ablation results of multi-level OIM losses on MARS are reported in Tab.~\ref{table: ablation of oim}.
The ``F-OIM'' denotes the frame-level OIM loss is deployed for each frame.
The ``V-OIM'' denotes the video-level OIM loss is utilized for each video.
%
%We can see that, the results show better retrieval accuracy when combining frame-level OIM loss and video-level OIM loss.
%
As shown in Tab.~\ref{table: ablation of oim}, higher performances are achieved when combining the frame-level and video-level OIM losses.
It demonstrates that multi-level losses could better optimize our proposed GRL.
%--------------------------------------------------------------------------------------
\subsection{Visualization Analysis}
We visualize the high-correlation maps and the accumulated low-correlation features in Fig.~\ref{fig:visualization}.
%
%At the second row of this figure, the high-correlation maps obtained from equation (2) are shown.
%
%It can be obviously observed that, the high-correlation maps will focus on the foreground regions.
%
Generally, features with high-correlations mean that they appear frequently in temporal and are spatially conspicuous.
Features with low-correlations mean that they are inconspicuous and discontinuous yet meaningful.
As shown in Fig.~\ref{fig:visualization}, the second row represents the high-correlation maps obtained from Equ.(2), covering the main and conspicuous regions, \emph{e.g.}, human upper body.
The third and forth rows show the accumulated low-correlation features, covering discontinuous but fine-grained cues, \emph{e.g.}, the bags or shoes.
Compared with the features learned from the high-correlation maps, the features from low-correlation maps in forward or backward process, could capture the incoherent and meaningful cues, such as shoes or bags, with red bounding boxes.
Meanwhile, we can find that, at the same time step, there are some variations among the features from the forward and backward process.
Thus, it is useful to assemble more discriminative information.
The visual maps further validate that our method could highlight the most conspicuous and aligned information in temporal and capture the sub-critical clues in spatial, simultaneously.
%---------------------------------------------------
\section{Conclusion}
In this paper, we propose a novel global-guided reciprocal learning framework for video-based person Re-ID.
%
%To capture multi-scale spatial relations of meaningful patterns,
We design a GCE module to estimate the correlation maps of spatial features under the global guidance.
Then, spatial features are disentangled into the high- and low-correlation features.
Besides, we propose a novel TRL module, in which multiple enhancement and memory units are designed for temporal learning.
%
%In a unit, two different strategies are adopted to enhance the high-correlation semantic features and accumulate more details from the low-correlation features.
%
%Besides, bi-directional process is arranged in forward and backward directions to resemble more robust representations.
%
Based on the proposed modules, our approach could not only enhance the conspicuous information from the high-correlation regions, but also accumulative fine-grained cues from the low-correlation features.
Extensive experiments on public benchmarks show that our framework outperforms several state-of-the-arts.

\textbf{Acknowledgements:} \small This work was supported in part by the National Key Research and Development Program of China under Grant No. 2018AAA0102001, the National Natural Science Foundation of China (NNSFC) under Grant No. 61725202, U1903215, 61829102, 91538201, 61771088,61751212, the Fundamental Research Funds for the Central Universities under Grant No. DUT20RC(3)083 and Dalian Innovation Leader’s Support Plan under Grant No. 2018RD07.

{\small
\bibliographystyle{ieee_fullname}
\bibliography{egbib}
}

\end{document}